# Real-Time Traffic Sign Detection: A Case Study in a Santa Clara Suburban Neighborhood

Harish Loghashankar[1], Hieu Nguyen[2]
[1]Student, Cupertino High School, Cupertino, California
[2]Mentor, University of Connecticut, Mansfield, Connecticut

ABSTRACT

This research project aims to develop a real-time traffic sign detection system using the YOLOv5 architecture and deploy it for efficient traffic sign recognition during a drive in a suburban neighborhood. The project's primary objectives are to train the YOLOv5 model on a diverse dataset of traffic sign images and deploy the model on a suitable hardware platform capable of real-time inference. The project will involve collecting a comprehensive dataset of traffic sign images. By leveraging the trained YOLOv5 model, the system will detect and classify traffic signs from a real-time camera on a dashboard inside a vehicle. The performance of the deployed system will be evaluated based on its accuracy in detecting traffic signs, real-time processing speed, and overall reliability. During a case study in a suburban neighborhood, the system demonstrated a notable 96% accuracy in detecting traffic signs. This research's findings have the potential to improve road safety and traffic management by providing timely and accurate real-time information about traffic signs and can pave the way for further research into autonomous driving.

## Introduction

According to the Annual United States Road Crash Statistics, more than 46,000 Americans die in car crashes yearly [1]. This disturbingly high number emphasizes the need for novel tools to help drivers on the road. With the recent rise of artificial intelligence and how quickly it has become mainstream in the status quo, utilizing these programs is one of the best ways to create tools to make roads safer. One such tool is accurate and timely traffic sign recognition, which can significantly assist drivers and reduce the number of fatal crashes each year. This proper recognition and prompt action in response to traffic signs is a challenge for drivers, as evidenced by 70,000 accidents occurring yearly just from failing to obey stop signs [2]. Real-time traffic sign detection using a camera inside the car can be a valuable tool for increasing road safety. It can instantaneously alert drivers about vital traffic signs on the road, such as stop signs, speed limits, or pedestrian crossings. This technology could also help prevent accidents caused by distracted drivers because real-time reminders and detections can help drivers stay focused on the road and aware of their surroundings [3]. In the future, such algorithms could be paired with autonomous vehicles to assist driving capabilities and prompt maneuvering to prevent a collision in response to road or traffic conditions. This work aims to generate a suitable algorithm to detect real-time traffic signage and test its efficacy and sensitivity. The approach to this work was completed using the YOLOv5 computer vision algorithm. The results of this work, along with the program having a 96% accuracy during the testing process in a suburban neighborhood, indicate that real-time traffic sign detection is feasible and can help make roads safer. The organization of this paper is as follows: the next section will cover the history of computer vision, an explanation of YOLOv5, followed by the dataset used for training the model, the process of training and testing the model, and finally addressing drawbacks that the model may have.

## A Brief History of Computer Vision

In the 1960s, researchers began to explore techniques for pattern recognition and image understanding with computers. [4] In 1966, researchers believed computer vision was so simple that they could create a functional computer vision model within just a summer by simply having a computer "describe what it [sees]" [5]. As they soon realized, computer vision was not that easy. During the 1970s, edge detection and feature extraction advancements occurred, laying the foundation for today's more sophisticated computer vision algorithms. [4] The decade after, researchers delved further into the quantitative side, a more mathematical view of computer vision. In the 1990s, object recognition, optical character recognition, and facial recognition were explored. Also, methods for 3-dimensional exact reconstructions of moments, just from images, were developed. In the late 1990s, sophisticated algorithms were invented, such as image-based rendering and image morphing. The early 2000s witnessed real-life machine learning applications, such as support vector machines and decision trees for object recognition tasks. [4] Figure 1 summarizes the progress of computer vision research.

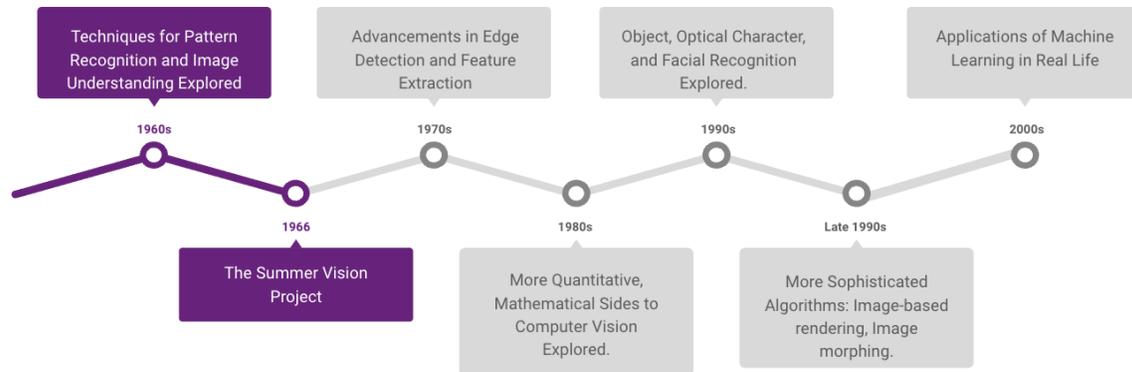

**Figure 1:** A Brief History of Computer Vision, in Timeline format.

In 2012, AlexNet [6], a deep CNN (convolutional neural network), achieved a breakthrough in image classification accuracy, igniting the deep learning revolution in computer vision. This revolution paved the way for the development of YOLO (You Only Look Once) [7] in 2016, which introduced real-time object detection by dividing images into a grid and predicting bounding boxes and class probabilities simultaneously in each grid. YOLO revolutionized object detection by significantly improving speed and accuracy, making it a widely embraced algorithm in various computer vision applications. The original YOLO model's best results only had a mAP (the mean of the accuracy and precision of the model) of 63.4%. It had a maximum of 45 fps while detecting objects in real-time on one of the best GPUs available after being trained on the PASCAL dataset. YOLOv2 had a higher mAP and fps at 76.8% and 67 fps, respectively, after being trained on the same dataset [8]. As newer YOLO versions were developed, the mAP percentages and fpses got higher even on more complex datasets like the COCO dataset, which is a large dataset dedicated to the purpose of object detection. It has over 330,000 images [9]. These reasons are why the YOLOv5 framework is an excellent option for traffic sign detection. It offers fast and accurate real-time performance, making it extremely useful for detecting traffic signs. The complete algorithm is open-sourced and is 3.6 gigabytes in size. Its ability to detect multiple objects in the same frame can help it see multiple traffic signs in complex traffic scenarios. YOLOv5 was used for this task instead of newer versions like YOLOv7 or YOLOv8 because YOLOv5 is integrated with PyTorch and is much more stable than more recent versions. The next section will go more in-depth about YOLOv5's architecture.

## Methodology - YOLOv5

YOLOv5 is an object recognition algorithm that can run in real-time. It can accurately identify and label desired objects in an image. Its architecture is shown below, in Figure 2. There are three stages to YOLOv5's detection algorithm. The first layer, the backbone, uses a modified CSPDarknet53 backbone, which extracts key features from the image it receives. This backbone is used because it solves the problem of duplicate gradient information, reducing complexity without losing accuracy [11]. The next layer, the neck, refines the features further. It uses PANet (Path Aggregation Network). PANet can better handle objects of different sizes in an image when compared to other convolutional neural networks. This makes it an ideal choice for traffic sign detection, as traffic signs are not all the same size. The final layer, the head, takes the refined features from the neck and uses them to generate predictions and bounding boxes. The head divides the image into cells and predicts the bounding boxes for objects inside each cell. After these predictions, an algorithm known as Non-Maximum Suppression (NMS) is used to remove repeated bounding boxes, making sure that each detected object is only labeled once. YOLOv5 has three types of loss functions, which it uses to train the model accurately. It has a classification loss function, which helps the model correctly predict the classes of objects. It also has a localization loss function, measuring the distances between the predicted bounding boxes and the actual bounding boxes. Finally, it has a confidence loss function, which helps generate the confidence scores for each prediction. With all these techniques put together, YOLOv5 can achieve incredible accuracy and precision in object detection.

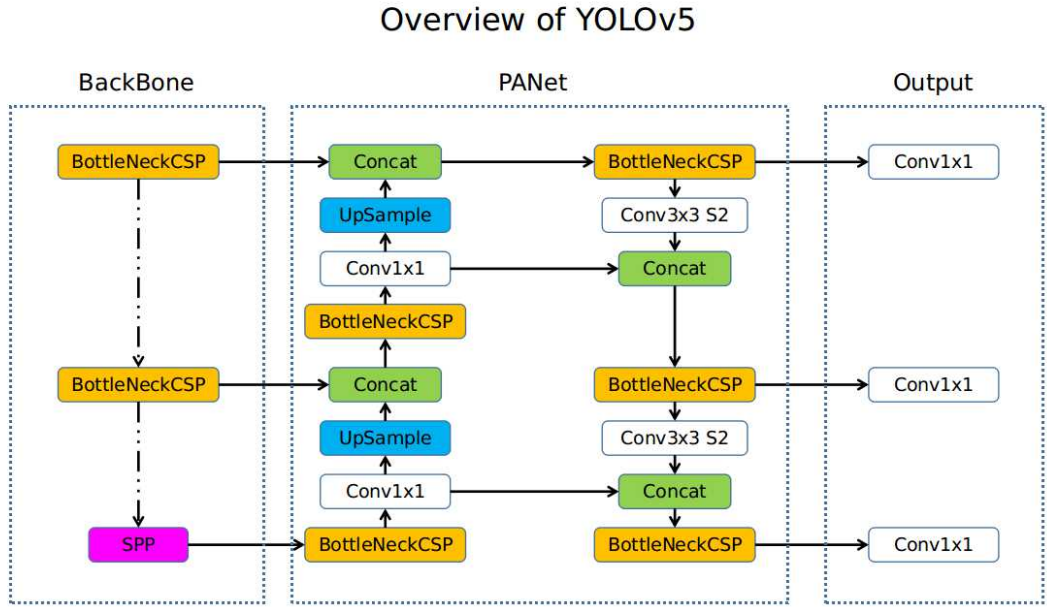

**Figure 2:** The YOLOv5 architecture, consisting of three parts: A Backbone, using CSPDarknet53, a Neck, using PANet, and a Head, outputting the predictions [13].

## Data

The dataset used for training and testing was the LISA (Laboratory for Intelligent Systems and Automation) dataset, a large publicly available dataset with annotated images of common American traffic signs that can be seen in everyday driving scenarios in the United States [10]. These common signs include stop signs, speed limits, school zones, and more. The images were annotated with bounding boxes, where each sign was identified by a labeled box around the sign. The images in the LISA dataset are incredibly diverse. Some photos are in black and white, while some are in color. The dataset also has diversity in environmental conditions, as it contains images in the rain and some with partially obscured signs. Given that the LISA dataset has many pictures, greater than 6,000, comprising of diverse environmental conditions, it is an excellent dataset for training a traffic sign detection model to improve overall accuracy. However, the dataset is incredibly skewed towards certain classes. The biasedness of the dataset could make the model biased in the same ways, like overpredicting for stop signs and underpredicting for do not enter signs. The figure below demonstrates the distribution of the LISA dataset.

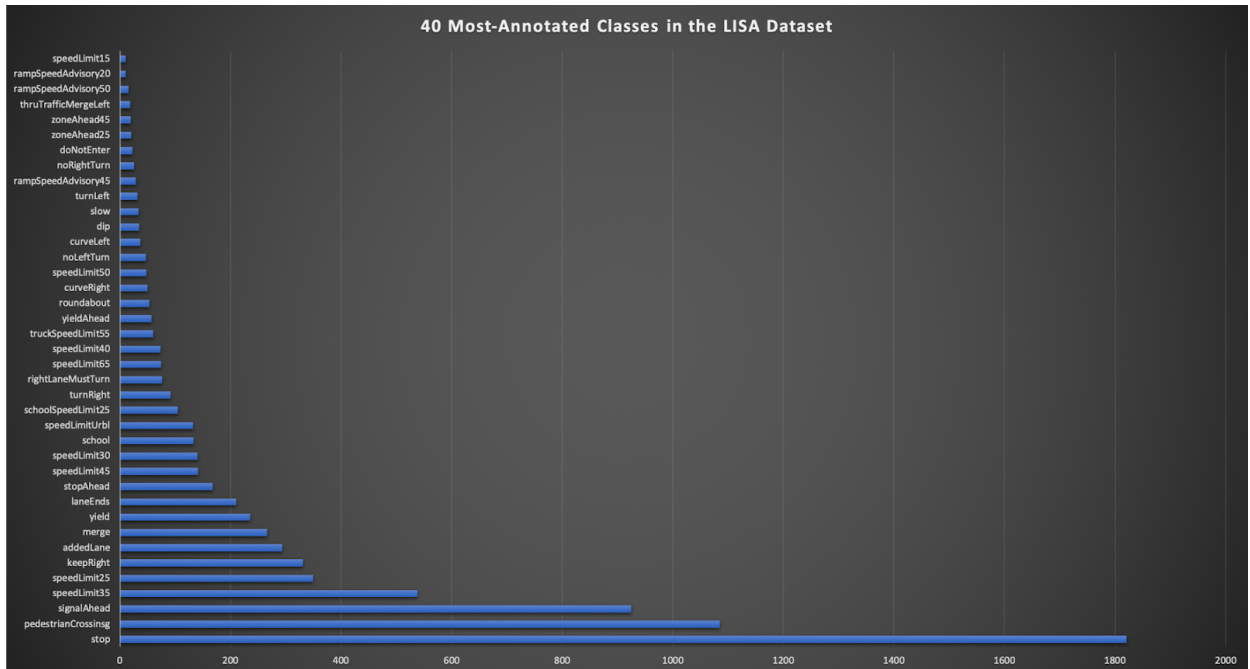

**Figure 3:** Distribution of Classes in the LISA Dataset. The stop sign class has the highest number of images included in the dataset, while speed limit 15 has one of the lowest.

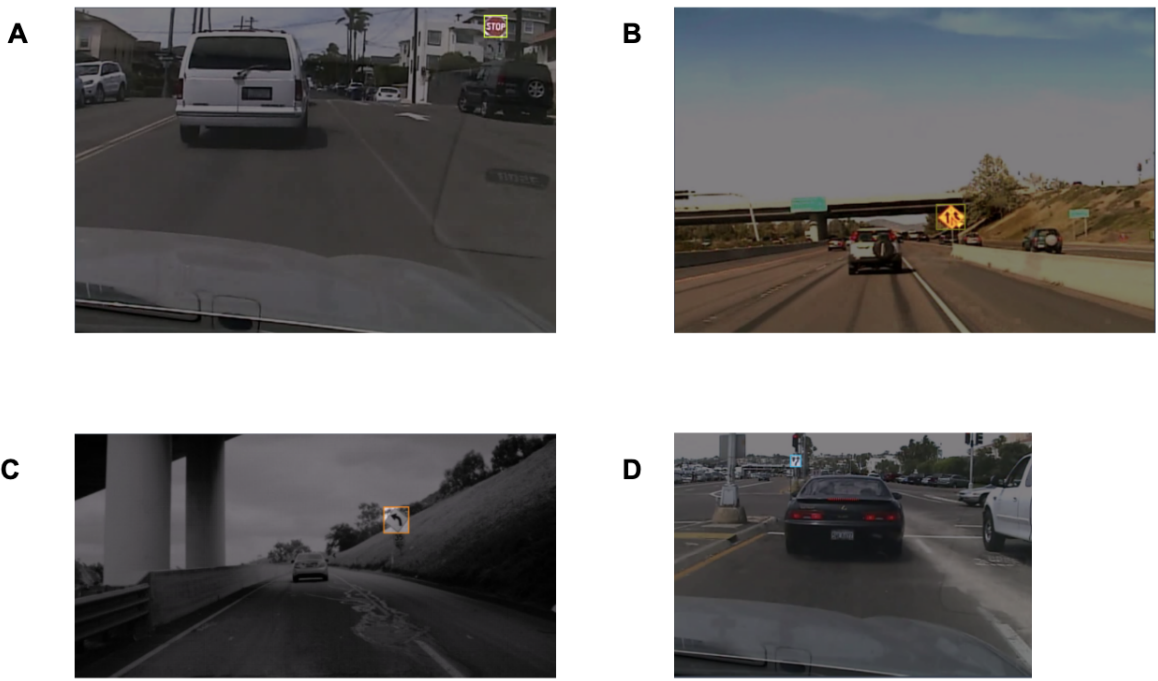

**Figure 4.** Examples of Bounding Box-Annotated Images from the LISA Dataset A: Image of a stop sign on a road, labeled with a bounding box. B: Image of a merge sign on a freeway, labeled with a bounding box. C: Image of a road curves left sign on a freeway ramp, labeled with a bounding box. D: image of a keep right sign on a road, labeled with a bounding box.

# Experiment

The training environment was set up by cloning the GitHub repository for YOLOv5 and installing the necessary modules and packages for it to function correctly. Then, the LISA dataset was labeled on Roboflow, with around 6,500 images, and was used for the training. Roboflow makes it easy to download datasets in whatever format is required. The dataset was downloaded in the YOLOv5 format. This was all done in Google Colab because it already contains a built-in CUDA (Compute Unified Device Architecture)-enabled Tesla T4 GPU. CUDA is a process that enables the speeding up of computer applications by utilizing the power of a system's GPUs. This means training could be completed, and the epochs would run faster on Google Colab's CUDA-enabled GPU than they ever could with an ordinary computer [12]. After the dataset had finished downloading, 132 epochs of training were run on Google Colab, and it took around a few hours to train thoroughly. Once done with training, it showed the model's mean accuracy precision (mAP). The average mAP for around 91% of the signs was more than 99%. However, due to the aforementioned skewing of the dataset, a few of the classes were severely underrepresented and had extremely low mAPs: For example, the "intersection" class had a final mAP of 0% as shown in Figure 5. This was because, as shown in Figure 3, the "intersection" class was not even in the top 40 most common signs.

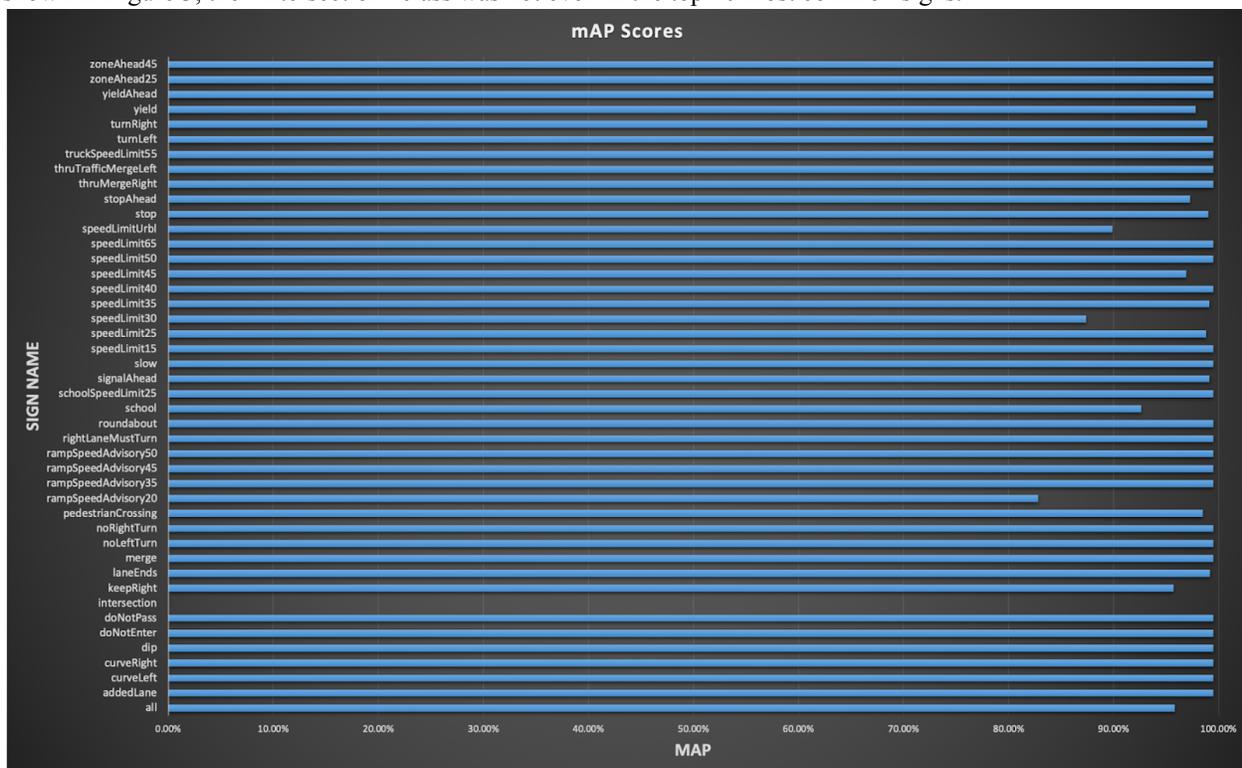

**Figure 5.** Post-training results: mAP indicates overall accuracy and precision. Most of the mAPs had a score of more than 90%, except for a few outliers; the underrepresented classes.

The weights were saved to a file inside the YOLOv5 directory when training was finished. Then, the weights could be downloaded and embedded into a local system for real-time testing. The next subsection goes into detail about a case study conducted with this model.

## Deployment: A Case Study of a Suburban Neighborhood

The detection algorithm ran in real-time during a drive around a Santa Clara suburban neighborhood in the afternoon during regular traffic. It was running for approximately 20 minutes, using the default laptop camera of a 2023 Macbook Air aimed toward the front of the car. Below are the results of how many traffic signs of each class were detected in the case study, and the overall performance of the model, and Figure 6 shows examples of successful detections by the model with high confidence.

| Sign Name | # of Signs detected correctly | # of Signs appearing in the drive |
|---|---|---|
| Pedestrian Crossing | 12 | 12 |
| Speed Limit 30 | 1 | 2 |
| Keep Right | 1 | 1 |
| School | 4 | 4 |
| Speed Limit 25 | 1 | 1 |
| Stop | 4 | 4 |
| Stop Ahead | 1 | 1 |
| Signal Ahead | 1 | 1 |
| Yield | 1 | 1 |
| **TOTAL** | **26** | **27** |

Table 1: Data collected from drive around neighborhood. 26 out of 27 times, the model was correct in its detection.

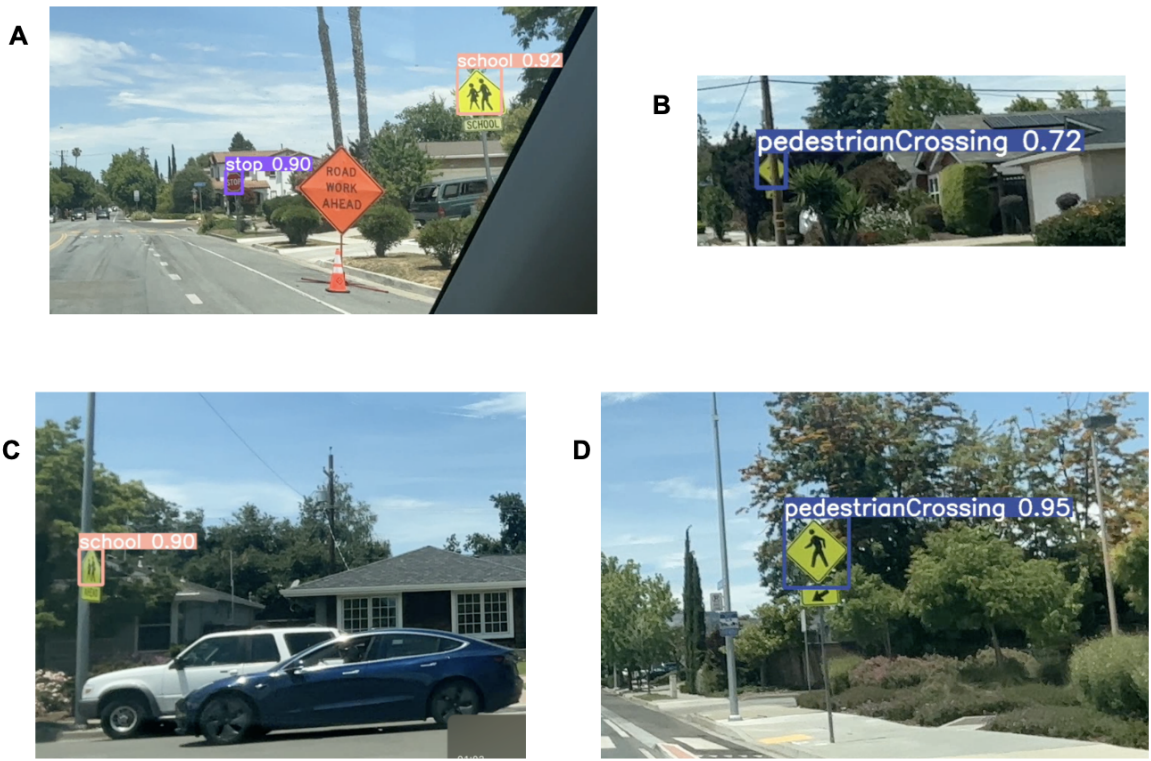

**Figure 6.** Images taken during drive around suburban neighborhood during the afternoon; A: The model detected both a stop sign and a school sign with high accuracies. B: The model detected a pedestrian crossing sign through a solid wooden pole. C: The model detected a school sign, even though it was slanted and barely readable. D: A normal detection of a pedestrian crossing sign.

The overall accuracy from all signs during this drive was 96%; the algorithm could detect most of the signs it was trained to. The model needed to be at most four car lengths away from the sign for accurate detection. This shows the potential real-world impacts of models like these.

## Addressing Drawbacks

Like all things, this model could have been improved in various ways. One drawback in the model stems from the training on the LISA dataset. This dataset's images were skewed, as shown earlier in the paper. Specifically, the stop sign class was significantly overrepresented, accounting for 1821 out of the 7855 total annotations, while other types were severely underrepresented, such as the intersection or speed limit 55 signs, accounting for just two images over the entire dataset. Subsequently, these classes' mAP was often <1% in these underrepresented sign classes and was rarely detected by the running algorithm. Below are some examples of incorrect detections made by the model during its real-time testing.

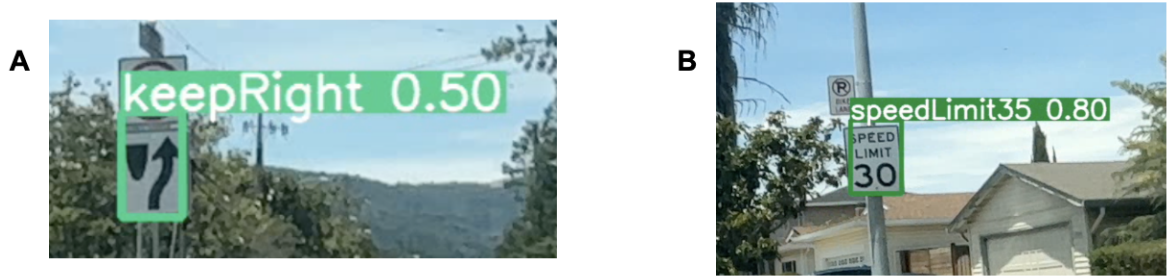

**Figure 7.** Examples of the imperfections of the model; low accuracy and incorrect detections. A: the model detected a "keepRight" sign correctly, but the model was only 50% sure that it was indeed a "keepRight" sign. B: the model incorrectly labels a "speedLimit30" sign as a "speedLimit35" sign.

To address this issue, a potential solution is to balance the dataset by uploading more annotated examples of these minority classes. This would substantially improve the model's accuracy and precision, as the model's bias would reduce and the model would detect signs more accurately.

Another drawback is the testing of the model in just one neighborhood. More rigorous experimentation would need to be done to test the limits of the environmental conditions the model was trained on with the LISA dataset to generate solid conclusions. For future research, more experimentation will be conducted on different neighborhoods, under different conditions, and with different camera settings.

## Conclusion

Now that artificial intelligence is being utilized in many daily tasks, the opportunities for making roads safer are immense. Using machine learning to train a traffic sign recognition model can guide drivers in decision-making and help prevent crashes due to missed information. The real-time accuracy rate of 96% shows how effective the model could be if deployed on a large scale; it could potentially assist millions of drivers in making better decisions on the road, ultimately resulting in fewer traffic accidents. This research could also be vital to building toward fully-automated self-driving cars; if these cars can read traffic signs, they should be able to navigate through most issues they face on the road. Though the model might not be perfect, drivers can take the information the model relays to them cautiously and should not rely on the model entirely.